\pdfoutput=1

\documentclass[10pt,twocolumn,letterpaper]{article}

\usepackage{cvpr}
\usepackage{times}
\usepackage{epsfig}
\usepackage{graphicx}
\usepackage{amsmath}
\usepackage{amssymb}
\usepackage[subrefformat=parens]{subcaption}
\usepackage{float}
\usepackage{afterpage}
\usepackage{comment}
\usepackage{caption}

\DeclareMathOperator{\sgn}{sgn}
\DeclareMathOperator*{\argmin}{argmin}

\cvprfinalcopy 

\def\cvprPaperID{5944} 

\begin{document}

\title{Inverse Path Tracing for Joint Material and Lighting Estimation}

\hypersetup{
	pdftitle={Inverse Path Tracing for Joint Material and Lighting Estimation},
	pdfauthor={Dejan Azinovi\'c, Tzu-Mao Li, Anton Kaplanyan and Matthias Nie{\ss}ner},
	pdfsubject={Computer Vision; Computer Graphics; Inverse Path Tracing; },
	pdfkeywords={Inverse Rendering, Path Tracing, Material Estimation, Lighting Estimation}}

\author{%
\hspace{-0.2cm}\parbox{4cm}{\centering Dejan Azinovi\'c$^{1}$}\quad
\parbox{4cm}{\centering Tzu-Mao Li$^{2,3}$}\quad
\parbox{4cm}{\centering Anton Kaplanyan$^{3}$}\quad
\parbox{4cm}{\centering Matthias Nie{\ss}ner$^{1}$}\\[0.8em]
 $^{1}$Technical University of Munich\quad
 $^{2}$MIT CSAIL\quad
 $^{3}$Facebook Reality Labs 
 \vspace{0.4cm}
}


\twocolumn[{%
\renewcommand\twocolumn[1][]{#1}%
\maketitle
\begin{center}
	\vspace{-0.9cm}
	\includegraphics[width=\linewidth]{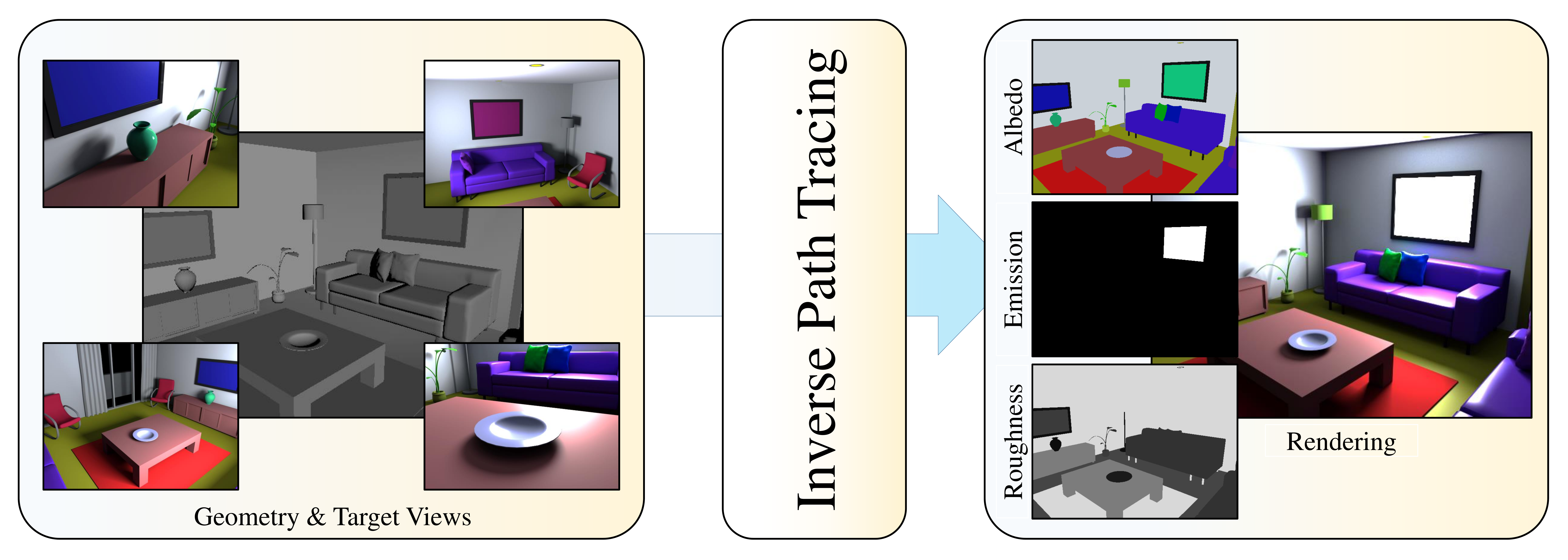}
	\vspace{-0.7cm}
    \captionof{figure}{
    Our Inverse Path Tracing algorithm takes as input a 3D scene and up to several RGB images (left), and estimates material as well as the lighting parameters of the scene. 
    The main contribution of our approach is the formulation of an end-to-end differentiable inverse Monte Carlo renderer which is utilized in a nested stochastic gradient descent optimization.
}
\label{fig:teaser}
\end{center}
}]

\newcommand{\x}{\mathbf{x}} 	
\newcommand{\y}{\mathbf{y}} 	
\newcommand{\n}{\mathbf{n}} 	
\newcommand{\wo}{\mathbf{i}}	
\newcommand{\wi}{\mathbf{o}}	
\newcommand{\X}{\mathbf{X}} 	



\begin{abstract}
Modern computer vision algorithms have brought significant advancement to 3D geometry reconstruction. 
However, illumination and material reconstruction remain less studied, with current approaches assuming very simplified models for materials and illumination.
We introduce Inverse Path Tracing, a novel approach to jointly estimate the material properties of objects and light sources in indoor scenes by using an invertible light transport simulation.
We assume a coarse geometry scan, along with corresponding images and camera poses.
The key contribution of this work is an accurate and simultaneous retrieval of light sources and physically based material properties (e.g., diffuse reflectance, specular reflectance, roughness, etc.) for the purpose of editing and re-rendering the scene under new conditions.
To this end, we introduce a novel optimization method using a differentiable Monte Carlo renderer that computes derivatives with respect to the estimated unknown illumination and material properties.
This enables joint optimization for physically correct light transport and material models using a tailored stochastic gradient descent. 
\end{abstract}


\section{Introduction}
With the availability of inexpensive, commodity RGB\mbox{-}D sensors, such as the Microsoft Kinect, Google Tango, or Intel RealSense, we have seen incredible advances in 3D reconstruction techniques~\cite{newcombe2011kinectfusion,izadi2011kinectfusion,niessner2013real,whelan2016elasticfusion,dai2017bundlefusion}.
While tracking and reconstruction quality have reached impressive levels, the estimation of lighting and materials has often been neglected.
Unfortunately, this presents a serious problem for virtual- and mixed-reality applications, where we need to re-render scenes from different viewpoints, place virtual objects, edit scenes, or enable telepresence scenarios where a person is placed in a different room.

This problem has been viewed in the 2D image domain, resulting in a large body of work on intrinsic images or videos \cite{Barron:2015:SIR,Meka:2016,Meka:2018:LLI}.
However, the problem is severely underconstrained on monocular RGB data due to lack of known geometry, and thus requires heavy regularization to jointly solve for lighting, material, and scene geometry.
We believe that the problem is much more tractable in the context of given 3D reconstructions. 
However, even with depth data available, most state-of-the-art methods, \eg, shading-based refinement~\cite{wu2014sfs,zollhoefer2015shading} or indoor re-lighting \cite{Zhang:2016:ERR}, are based on simplistic lighting models, such as spherical harmonics (SH)~\cite{Ramamoorthi:2001:SFI} or spatially-varying SH~\cite{maier2017intrinsic3d}, which can cause issues on occlusion and view-dependent effects (Fig.~\ref{fig:shadow}).

In this work, we address this shortcoming by formulating material and lighting estimation as a proper inverse rendering problem.
To this end, we propose an Inverse Path Tracing algorithm that takes as input a given 3D scene along with a single or up to several captured RGB frames.
The key to our approach is a differentiable Monte Carlo path tracer which can differentiate with respect to rendering parameters constrained on the difference of the rendered image and the target observation.
Leveraging these derivatives, we solve for the material and lighting parameters by nesting the Monte Carlo path tracing process into a stochastic gradient descent (SGD) optimization.
The main contribution of this work lies in this SGD optimization formulation, which is inspired by recent advances in deep neural networks.

\begin{figure}[thb!]
\begin{center}
\vspace{-0.4cm}
\includegraphics[width=\linewidth]{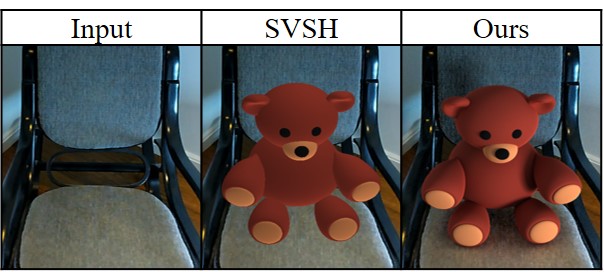}
\end{center}
\vspace{-0.6cm}
   \caption{Inserting virtual objects in real 3D scenes; the estimated lighting and material parameters of our approach enable convincing image compositing in AR settings.}
   \vspace{-0.2cm}
\label{fig:matterport_teddy}
\end{figure}

We tailor this Inverse Path Tracing algorithm to 3D scenes, where scene geometry is (mostly) given but the material and lighting parameters are unknown.
In a series of experiments on both synthetic ground truth and real scan data, we evaluate the design choices of our optimizer.
In comparison to current state-of-the-art lighting models, we show that our inverse rendering formulation and its optimization achieves significantly more accurate results. 

In summary, we contribute the following:

\begin{itemize}
\item \vspace{-0.2cm}
An end-to-end differentiable inverse path tracing formulation for joint material and lighting estimation.
\item \vspace{-0.2cm}
A flexible stochastic optimization framework with extensibility and flexibility for different materials and regularization terms.
\end{itemize}

\section{Related Work}

Material and illumination reconstruction has a long history in computer vision (\eg,~\cite{Patow:2003:SIR, Bonneel:IDI:2017}).
Given scene geometry and observed radiance of the surfaces, the task is to infer the material properties and locate the light source.
However, to our knowledge, none of the existing methods handle non-Lambertian materials with near-field illumination (area light sources), while taking interreflection between surfaces into account.

\textbf{3D approaches.} A common assumption in reconstructing material and illumination is that the light sources are infinitely far away.
Ramamoorthi and Hanrahan~\cite{Ramamoorthi:2001:SFI} project both material and illumination onto spherical harmonics and solve for their coefficients using the convolution theorem. 
Dong~\etal~\cite{Dong:2014:ARS} solve for spatially-varying reflectance from a video of an object. 
Kim~\etal~\cite{Kim:2017:LAO} reconstruct the reflectance by training a convolutional neural network operating on voxels constructed from RGB-D video.
Maier~\etal~\cite{maier2017intrinsic3d} generalize spherical harmonics to handle spatial dependent effects, but do not correctly take view-dependent reflection and occlusion into account.
All these approaches simplify the problem by assuming that the light sources are infinitely far away, in order to reconstruct a single environment map shared by all shading points. 
In contrast, we model the illumination as emission from the surfaces, and handle near-field effects such as the squared distance falloff or glossy reflection better.

\textbf{Image-space approaches (\eg,~\cite{Barrow:1978:RIS, Barron:2015:SIR, Deschaintre:2018:SSC, Meka:2018:LLI}).} 
These methods usually employ sophisticated data-driven approaches, by learning the distributions of material and illumination. 
However, these methods do not have a notion of 3D geometry, and cannot handle occlusion, interreflection and geometry factors such as the squared distance falloff in a physically based manner. 
These methods also usually require a huge amount of training data, and are prone to errors when subjected to scenes with different characteristics from the training data.

\textbf{Active illumination (\eg,~\cite{Marschner:1998:IRC, Debevec:2000:ARF, Kang:2018:ERC}).} 
These methods use highly-controlled lighting for reconstruction, by carefully placing the light sources and measuring the intensity. 
These methods produce high-quality results, at the cost of a more complicated setup.

\textbf{Inverse radiosity (\eg,~\cite{Yu:1999:IGI, Zhang:2016:ERR})} achieves impressive results for solving near-field illumination and Lambertian materials for indoor illumination. 
It is difficult to generalize the radiosity algorithm to handle non-Lambertian materials (Yu~\etal handle it by explicitly measuring the materials, whereas Zhang~\etal assume Lambertian). 

\textbf{Differentiable rendering.} Blanz and Vetter utilized differentiable rendering for face reconstruction using 3D morphable models~\cite{Blanz:1999:MMS}, which is now leveraged by modern analysis-by-synthesis face trackers~\cite{thies2016face2face}.
Gkioulekas~\etal~\cite{Gkioulekas:2013:IVR,Gkioulekas:2016:ECI} and Che~\etal~\cite{Che:2018:ITN} solve for scattering parameters using a differentiable volumetric path tracer.
Kasper~\etal~\cite{Kasper:2017:LSE} developed a differentiable path tracer, but focused on distant illumination.
Loper and Black~\cite{Loper:2014:OAD} and Kato~\cite{Kato:2018:N3M} developed fast differentiable rasterizers, but do not support global illumination.
Li~\etal~\cite{Li:2018:DMC} showed that it is possible to compute correct gradients of a path tracer while taking discontinuities introduced by visibility into consideration.

\section{Method}
\afterpage{\begin{figure*}[ht]
  \centering
  \captionsetup[subfigure]{justification=centering}
  \begin{subfigure}{0.19\textwidth}
  {\includegraphics[width=0.99\linewidth]{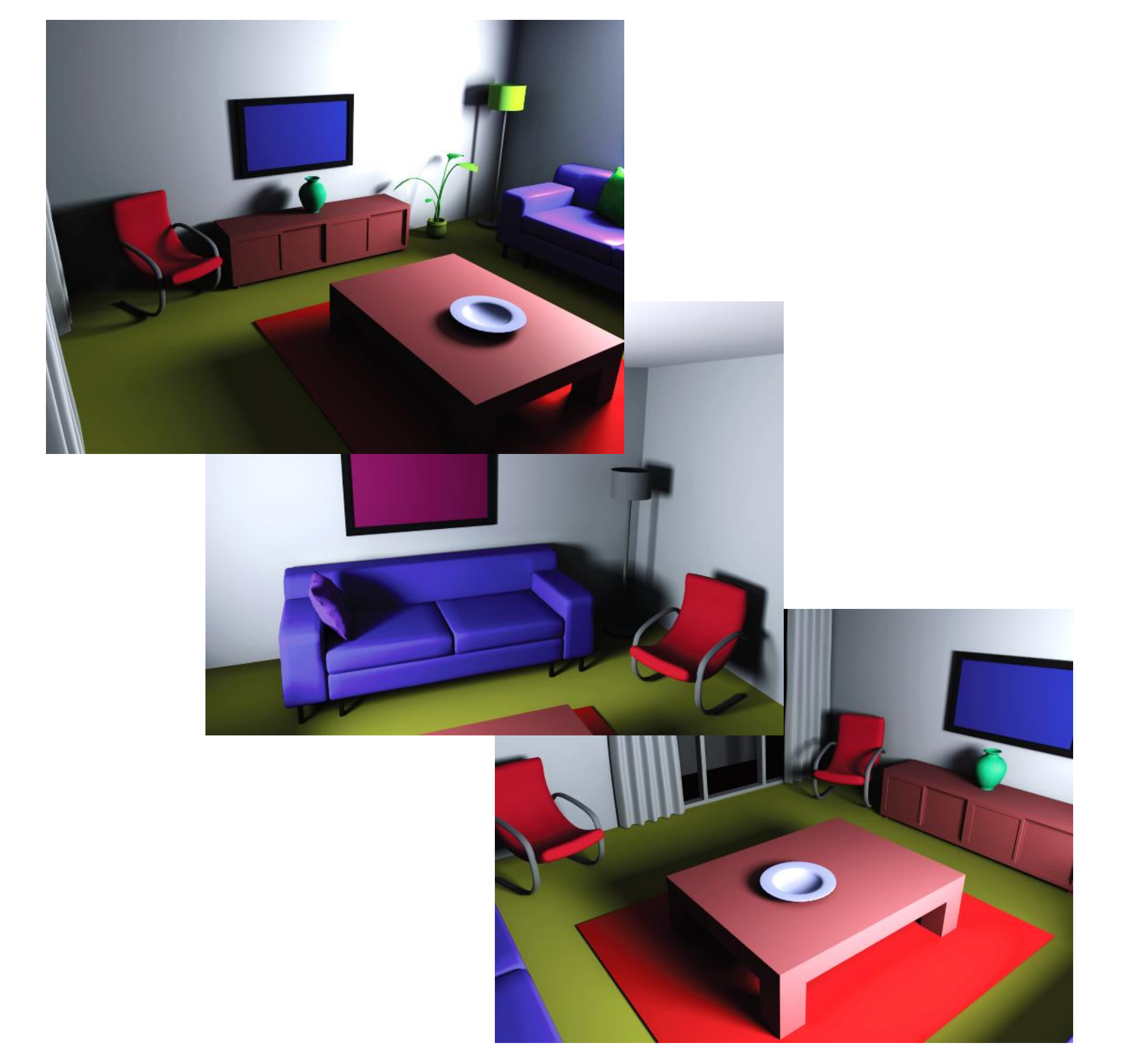} \vspace{-0.3cm}}
  \caption{input photos}
  \label{fig:input_photos}
  \end{subfigure}
  \begin{subfigure}{0.19\textwidth}
  {\includegraphics[width=0.99\linewidth]{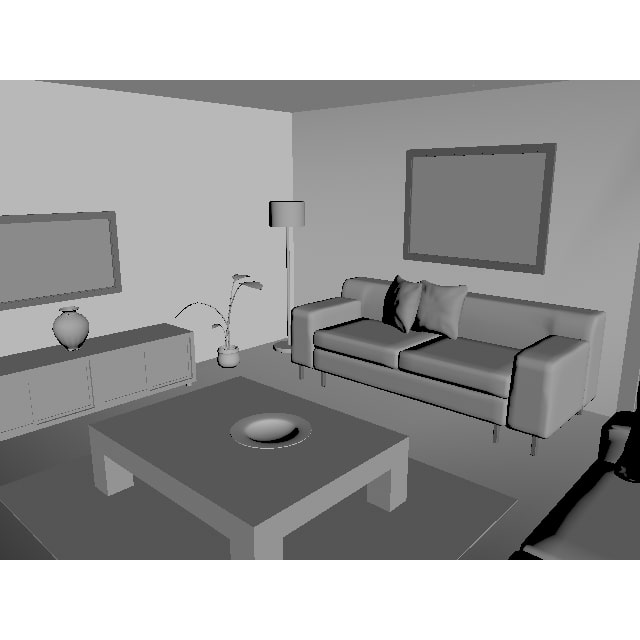} \vspace{-0.3cm}}
  \caption{geometry scan \& \\
           object segmentation}
  \label{fig:geometry_scan}
  \end{subfigure}
  \begin{subfigure}{0.19\textwidth}
  {\includegraphics[width=0.99\linewidth]{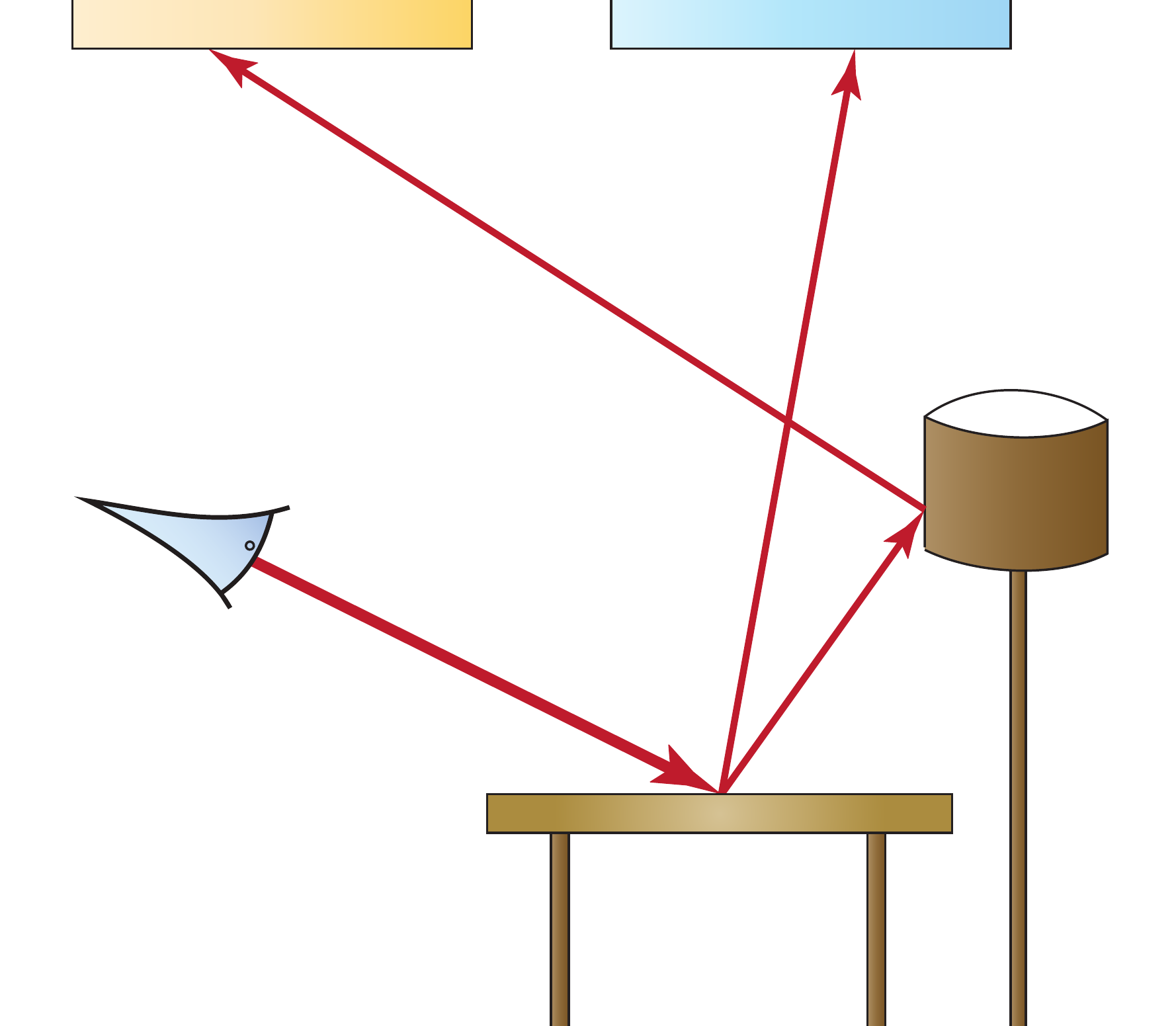} \vspace{-0.3cm}}
  \caption{path tracing}
  \label{fig:path_tracing}
  \end{subfigure}
  \begin{subfigure}{0.19\textwidth}
  {\includegraphics[width=0.99\linewidth]{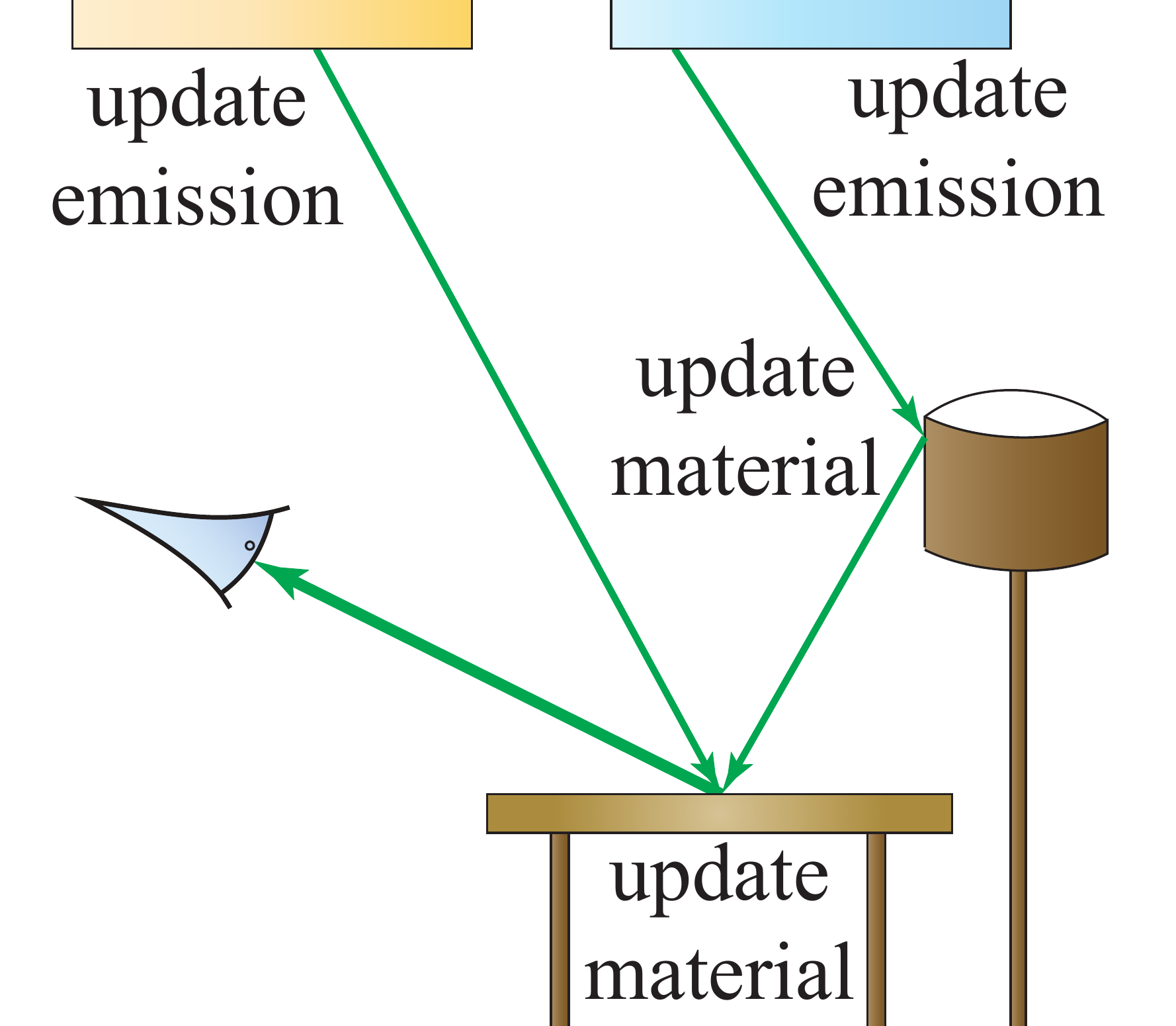} \vspace{-0.3cm}}
  \caption{backpropagate}
  \label{fig:path_tracing_backprop}
  \end{subfigure}
  \begin{subfigure}{0.19\textwidth}
  {\includegraphics[width=0.99\linewidth]{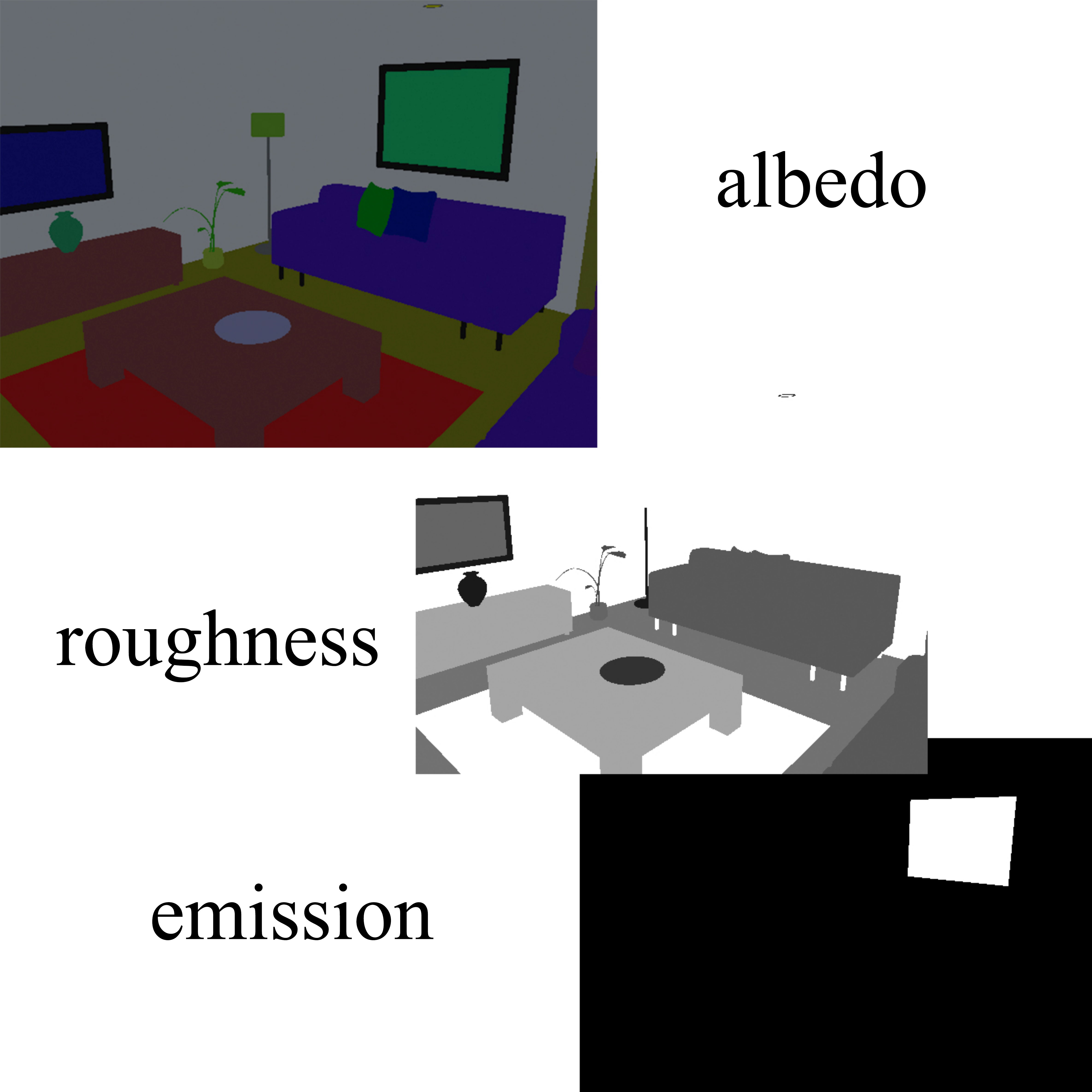} \vspace{-0.3cm}}
  \caption{reconstructed \\ materials \& illumination}
  \label{fig:reconstructed}
  \end{subfigure}
\caption{Overview of our pipeline. Given (a) a set of input photos from different views, along with (b) an accurate geometry scan and proper segmentation, we reconstruct the material properties and illumination of the scene, by iteratively (c) rendering the scene with path tracing, and (d) backpropagating to the material and illumination parameters in order to update them. After numerous iterations, we obtain the (e) reconstructed material and illumination.}
\label{fig:overview}
\end{figure*}}

Our \emph{Inverse Path Tracing} method employs physically based light transport simulation~\cite{Kajiya86} to estimate derivatives of all unknown parameters \wrt the rendered image(s). 
The rendering problem is generally extremely high-dimensional and is therefore usually solved using stochastic integration methods, such as Monte Carlo integration. 
In this work, we nest differentiable path tracing into stochastic gradient descent to solve for the unknown scene parameters.
Fig.~\ref{fig:overview} illustrates the workflow of our approach. 
We start from the captured imagery, scene geometry, object segmentation of the scene, and an arbitrary initial guess of the illumination and material parameters.
Material and emission properties are then estimated by optimizing for rendered imagery to match the captured images.

The path tracer renders a noisy and undersampled version of the image using Monte Carlo integration and computes derivatives of each sampled light path \wrt the unknowns. 
These derivatives are passed as input to our optimizer to perform a single optimization step. 
This process is performed iteratively until we arrive at the correct solution. 
Path tracing is a computationally expensive operation, and this optimization problem is non-convex and ill-posed. 
To this end, we employ variance reduction and novel regularization techniques (Sec.~\ref{ssec:regularization}) for our gradient computation to arrive at a converged solution within a reasonable amount of time, usually a few minutes on a modern 8-core CPU.

\subsection{Light Transport Simulation}
\label{sec:pt}
If all scene and image parameters are known, an expected linear pixel intensity can be computed using light transport simulation. In this work, we assume that all surfaces are opaque and there is no participating media (\eg, fog) in the scene. In this case, the rendered intensity  $I_R^j$ for pixel $j$ is computed using the path integral~\cite{Veach98}:
\begin{align}
	I_{R}^{j} = \int_{\Omega} h_j(\X) f(\X) d \mu(\X),
      \label{eq:path_integral}
\end{align}
where $\X=(\x_0,...,\x_k)$ is a light path, \ie a list of vertices on the surfaces of the scene starting at the light source and ending at the sensor; the integral is a path integral taken over the space of all possible light paths of all lengths, denoted as $\Omega$, with a product area measure $\mu(\cdot)$; $f(\X)$ is the measurement
contribution function of a light path $\X$ that computes how much energy flows through this particular path;
and $h_j(\X)$ is the pixel filter kernel of the sensor's pixel $j$, which is non-zero only when the light path $\X$ ends around the pixel $j$ and incorporates sensor sensitivity at this pixel.
We refer interested readers to the work of Veach~\cite{Veach98} for more details on the light transport path integration.

The most important term of the integrand to our task is the path measurement contribution function $f$, as it contains the material parameters as well as the information about the light sources. For a path $\X=(\x_0,...,\x_k)$ of length $k$, the measurement contribution function has the following form:
\begin{align}
	f(\X) = L_{\text{e}}(\x_0, \overline{\x_0\x_1}) 
      \prod_{i=1}^{k} f_{\text{r}}(\x_i, \,\overline{\x_{i\!-\!1}\x_i}, \,\overline{\x_i \x_{i+\!1}}), 
    \label{eq:meas}
\end{align}
where $L_{\text{e}}$ is the radiance emitted at the scene surface point $\x_0$ (beginning of the light path) towards the direction $\overline{\x_0\x_1}$. 
At every interaction vertex $\x_i$ of the light path, there is a \emph{bidirectional reflectance distribution function} (BRDF) $f_r(\x_i, \overline{\x_{i\!-\!1}\x_i}, \overline{\x_i \x_{i+\!1}})$ defined.
The BRDF describes the material properties at the point $\x_i$, \ie, how much light is scattered from the incident direction $\overline{\x_{i\!-\!1}\x_i}$ towards the outgoing direction $\overline{\x_i \x_{i+\!1}}$. 
The choice of the parametric BRDF model $f_r$ is crucial to the range of materials that can be reconstructed by our system. We discuss the challenges of selecting the BRDF model in Sec.~\ref{sec:params}.

Note that both the BRDF $f_r$ and the emitted radiance $L_e$ are unknown and the desired parameters to be found at every point on the scene manifold.

\subsection{Optimizing for Illumination and Materials}
We take as input a series of images in the form of real-world photographs or synthetic renderings, together with the reconstructed scene geometry and corresponding camera poses. We aim to solve for the unknown material parameters $\mathcal{M}$ and lighting parameters $\mathcal{L}$ that will produce rendered images of the scene that are identical to the input images. 

Given the un-tonemapped captured pixel intensities $I_C^j$ at all pixels $j$ of all images, and the corresponding noisy estimated pixel intensities $\tilde{I}_{R}^j$ (in linear color space), we seek all material and illumination parameters $\Theta=\{\mathcal{M}, \mathcal{L}\}$ by solving the following optimization problem using stochastic gradient descent:
\begin{align}
	\argmin_\Theta E(\Theta) = \sum_{j}^{N} \left|I_C^{j} - \tilde{I}_{R}^{j}\right|_1,
    \label{eq:sgd}
\end{align}
where $N$ is the number of pixels in all images. 
We found that using an $L_1$ norm as a loss function helps with robustness to outliers, such as extremely high contribution samples coming from Monte Carlo sampling. 

\subsection{Computing Gradients with Path Tracing}
In order to efficiently solve the minimization problem in Eq.~\ref{eq:sgd} using stochastic optimization, we compute the gradient of the energy function $E(\Theta)$ with respect to the set of unknown material and emission parameters $\Theta$:
\begin{align}
	\nabla_\Theta E(\Theta) = 
    \sum_{j}^{N} \nabla_\Theta \tilde{I}_{R}^{j}    \sgn\left(I_C^{j} - \tilde{I}_{R}^{j}\right),
    \label{eq:sgd_grad}
\end{align}
where $\sgn(\cdot)$ is the sign function, and $\nabla_\Theta \tilde{I}_{R}^{j}$ the gradient of the Monte Carlo estimate with respect to all unknowns $\Theta$.

Note that this equation for computing the gradient now has two Monte Carlo estimates for each pixel $j$: (1) the estimate of pixel color itself $\tilde{I}_{R}^{j}$; and (2) the estimate of its gradient $\nabla_\Theta \tilde{I}_{R}^{j}$. 
Since the expectation of product only equals the product of expectation when the random variables are independent, it is important to draw independent samples for each of these estimates to avoid introducing bias.

In order to compute the gradients of a Monte Carlo estimate for a single pixel $j$, we determine what unknowns are touched by the measurement contribution function $f(\X)$ for a sampled light path $\X$. 
We obtain the explicit formula of the gradients by differentiating Eq.~\ref{eq:meas} using the product rule (for brevity, we omit some arguments for emission $L_{\text{e}}$ and BRDF $f_{\text{r}}$):
\begin{align}
	\nabla_{\Theta_\mathcal{L}} f(\X) &= \nabla_{\Theta_\mathcal{L}}L_{\text{e}}(\x_0)
      \prod_{i}^{k} f_{\text{r}}(\x_i)
\label{eq:emission_gradients}
      \\
	\nabla_{\Theta_\mathcal{M}} f(\X) &= L_{\text{e}}(\x_0) \sum_l^k 
      \nabla_{\Theta_\mathcal{M}}f_{\text{r}}(\x_l) \!\!\prod_{i, i \neq l}^{k} \!\!f_{\text{r}}(\x_i) 
\label{eq:material_gradients}
\end{align}
where the gradient vector $\nabla_{\Theta} = \{\nabla_{\Theta_\mathcal{M}}, \nabla_{\Theta_\mathcal{L}}\}$ is very sparse and has non-zero values only for unknowns touched by the path $\X$.
The gradients of emissions (Eq.~\ref{eq:emission_gradients}) and materials (Eq.~\ref{eq:material_gradients}) have similar structure to the original path contribution (Eq.~\ref{eq:meas}). 
Therefore, it is natural to apply the same path sampling strategy; see the appendix for details.

\subsection{Multiple Captured Images}
The single-image problem can be directly extended to multiple images. 
Given multiple views of a scene, we aim to find parameters for which rendered images from these views match the input images. 
A set of multiple views can cover parts of the scene that are not covered by any single view from the set. 
This proves important for deducing the correct position of the light source in the scene. 
With many views, the method can better handle view-dependent effects such as specular and glossy highlights, which can be ill-posed with just a single view, as they can also be explained as variations of albedo texture.

\section{Optimization Parameters and Methodology}

In this section we address the remaining challenges of the optimization task: what are the material and illumination parameters we actually optimize for, and how to resolve the ill-posed nature of the problem.

\subsection{Parametric Material Model}
\label{sec:params}

We want our material model to satisfy several properties.
First, it should cover as much variability in appearance as possible, including such common effects as specular highlights, multi-layered materials, and spatially-varying textures.
On the other hand, since each parameter adds another unknown to the optimization, we would like to keep the number of parameters minimal.
Since we are interested in re-rendering and related tasks, the material model needs to have interpretable parameters, so the users can adjust the parameters to achieve the desired appearance.
Finally, since we are optimizing the material properties using first-order gradient-based optimization, we would like the range of the material parameters to be similar.

To satisfy these properties, we represent our materials using the Disney material model~\cite{burley2012physically}, the state-of-the-art physically based material model used in movie and game rendering. 
It has a ``base color'' parameter which is used by both diffuse and specular reflectance, as well as $10$ other parameters describing the roughness, anisotropy, and specularity of the material. All these parameters are perceptually mapped to $[0, 1]$, which is both interpretable and suitable for optimization.

\begin{figure}[htb!]
\begin{center}
\includegraphics[width=\linewidth]{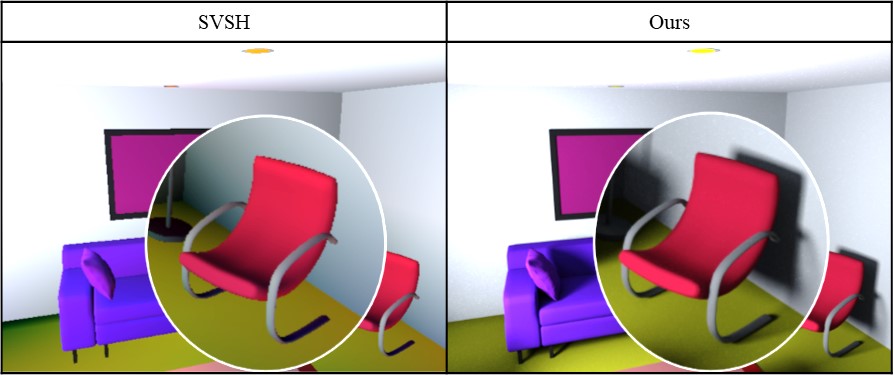}
\end{center}
\vspace{-0.5cm}
   \caption{Methods based on spherical harmonics have difficulties handling sharp shadows or lighting changes due to the distant illumination assumption. A physically based method, such as Inverse Path Tracing,  correctly reproduces these effects.}
\vspace{-0.3cm}
\label{fig:shadow}
\end{figure}

\subsection{Scene Parameterization}
We use triangle meshes to represent the scene geometry. 
Surface normals are defined per-vertex and interpolated within each triangle using barycentric coordinates. The optimization is performed on a per-object basis, \ie, every object has a single unknown emission and a set of material parameters that are assumed constant across the whole object. 
We show that this is enough to obtain accurate lighting and an average constant value for the albedo of an object.

\subsection{Emission Parameterization}
For emission reconstruction, we currently assume all light sources are scene surfaces with an existing reconstructed geometry. 
For each emissive surface, we currently assume that emitted radiance is distributed according to a view-independent directional emission profile $L_e(\x,\wo)=e(\x) (\wo\cdot\n(\x))_+$, where $e(\x)$ is the unknown radiant flux at $\x$; $\wo$ is the emission direction at surface point $\x$, $\n(\x)$ is the surface normal at $\x$ and $(\cdot)_+$ is the dot product (cosine) clamped to only positive values. 
This is a common emission profile for most of the area lights, which approximates most of the real soft interior lighting well. 
Our method can also be extended to more complex or even unknown directional emission profiles or purely directional distant illumination (\eg, sky dome, sun) if needed.

\subsection{Regularization}
\label{ssec:regularization}
The observed color of an object in a scene is most easily explained by assigning emission to the triangle. 
This is only avoided by differences in shading of the different parts of the object. 
However, it can happen that there are no observable differences in the shading of an object, especially if the object covers only a few pixels in the input image. 
This can be a source of error during optimization. 
Another source of error is Monte Carlo and SGD noise. 
These errors lead to incorrect emission parameters for many objects after the optimization. 
The objects usually have a small estimated emission value when they should have none. We tackle the problem with an L1-regularizer for the emission. 
The vast majority of objects in the scene is not an emitter and having such a regularizer suppresses the small errors we get for the emission parameters after optimization.

\subsection{Optimization Parameters}
We use ADAM~\cite{Kingma:2015:AMS} as our optimizer with batch size $B=8$ estimated pixels and learning rate $5\cdot10^{-3}$. 
To form a batch, we sample $B$ pixels uniformly from the set of all pixels of all images.
Please see the appendix for an evaluation regarding the impact of different batch sizes and sampling distributions on the convergence rate.
While a higher batch size reduces the variance of each iteration, having smaller batch sizes, and therefore faster iterations, proves to be more beneficial.
\begin{figure*}[htb!]
\begin{center}
\includegraphics[width=0.99\linewidth]{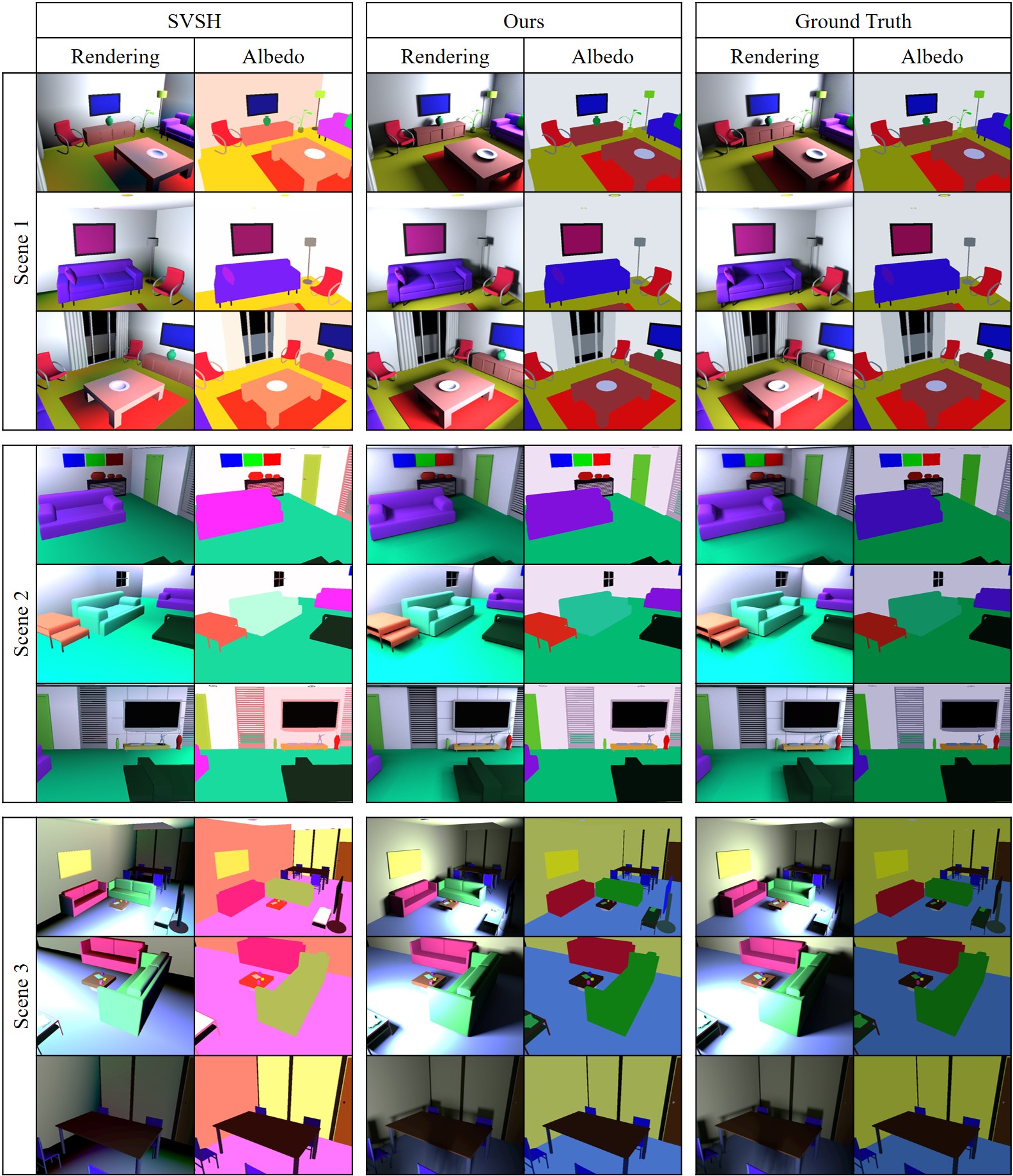}
\end{center}
   \caption{Evaluation on synthetic scenes. Three scenes have been rendered
   from different views with both direct and indirect lighting (right).
   An approximation of the albedo lighting with spatially-varying spherical
   harmonics is shown (left). Our method is able to detect the light
   source even though it was not observed in any of the views (middle). Notice 
   that we are able to reproduce sharp lighting changes and shadows correctly. The 
   albedo is also closer to the ground truth albedo.}
\label{fig:synth}
\end{figure*}
\section{Results}
\label{sec:results}

\begin{figure}[htb!]
\begin{center}
\includegraphics[width=\linewidth]{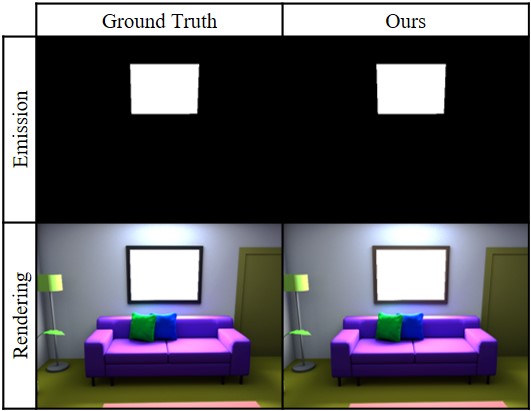}
\end{center}
\vspace{-0.5cm}
   \caption{Inverse Path Tracing is able to correctly detect the light emitting object (top). The ground truth rendering and our estimate is shown on the bottom. Note that this view was not used during optimization.}
\label{fig:emission}
\end{figure}

\begin{figure}[htb!]
\begin{center}
\includegraphics[width=\linewidth]{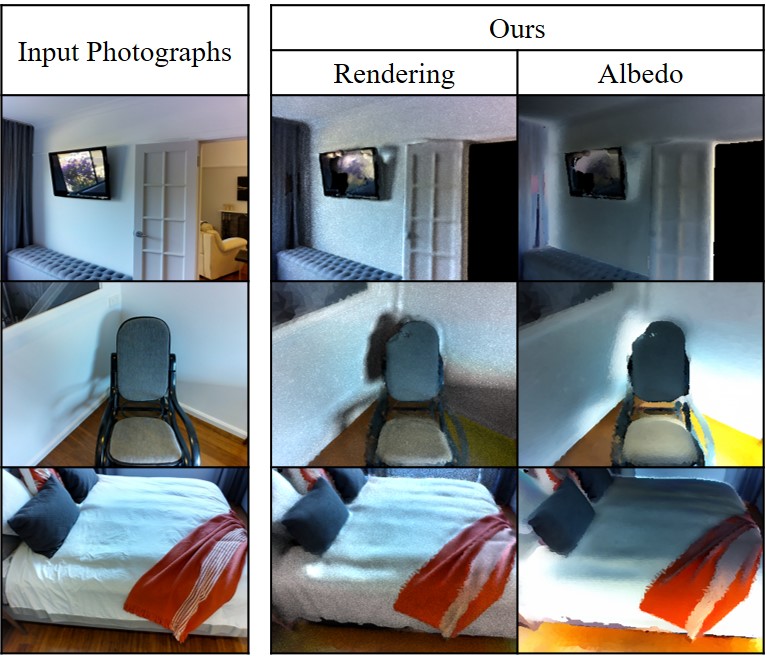}
\end{center}
\vspace{-0.5cm}
   \caption{We can resolve object textures by optimizing for the unknown parameters per triangle. Higher resolution textures can be obtained by further subdividing the geometry.}
\label{fig:matterport_subdivision}
\end{figure}

\paragraph{Evaluation on synthetic data.} We first evaluate our method
on multiple synthetic scenes, where we know the ground truth solution. Quantitative results are listed in Tab.~\ref{tab:quantitative}, and qualitative results are shown in Fig.~\ref{fig:synth}. Each scene is rendered using a path tracer with the ground truth lighting and materials to obtain the ``captured images''. These captured images and scene geometry are then given to our Inverse Path Tracing algorithm, which optimizes for unknown lighting and material parameters. We compare to the closest  previous work based on spatially-varying spherical harmonics (SVSH)~\cite{maier2017intrinsic3d}. SVSH fails to capture sharp details such as shadows or high-frequency lighting changes. A comparison of the shadow quality is presented in Fig.~\ref{fig:shadow}.

Our method correctly detects light sources and converges to a correct emission value, while the emission of objects that do not emit light stays at zero. Fig.~\ref{fig:emission} shows a novel view, rendered with results from an optimization that was performed on input views from Fig.~\ref{fig:synth}. Even though the light source was not visible in any of the input views, its emission was correctly computed by Inverse Path Tracing.

\begin{figure}[htb!]
\begin{center}
\includegraphics[width=\linewidth]{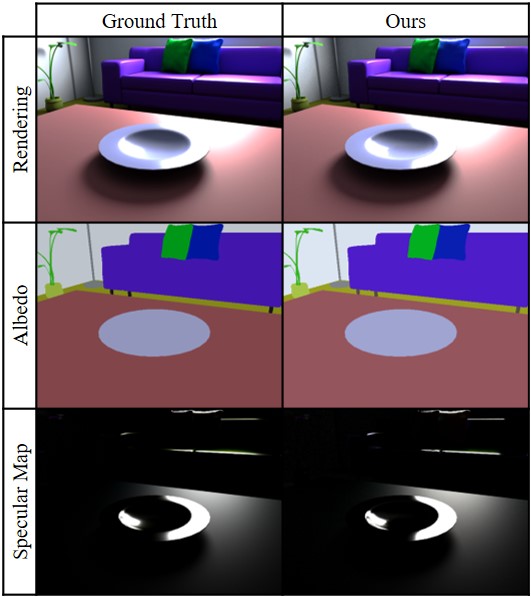}
\end{center}
\vspace{-0.5cm}
   \caption{Inverse Path Tracing is agnostic to the underlying BRDF; \eg, here, in a specular case, we are able to correctly estimate both the albedo and the roughness of the objects. The ground truth rendering and our estimate is shown on top, the albedo in the middle and the specular map on the bottom.}
\label{fig:specular}
\end{figure}

In addition to albedo, our Inverse Path Tracer can also optimize for other material parameters such as roughness. 
In Fig.~\ref{fig:specular}, we render a scene containing objects of varying roughness. Even when presented with the challenge of estimating both albedo and roughness, our method produces the correct result as shown in the re-rendered image.

\begin{figure}[htb!]
\begin{center}
\includegraphics[width=\linewidth]{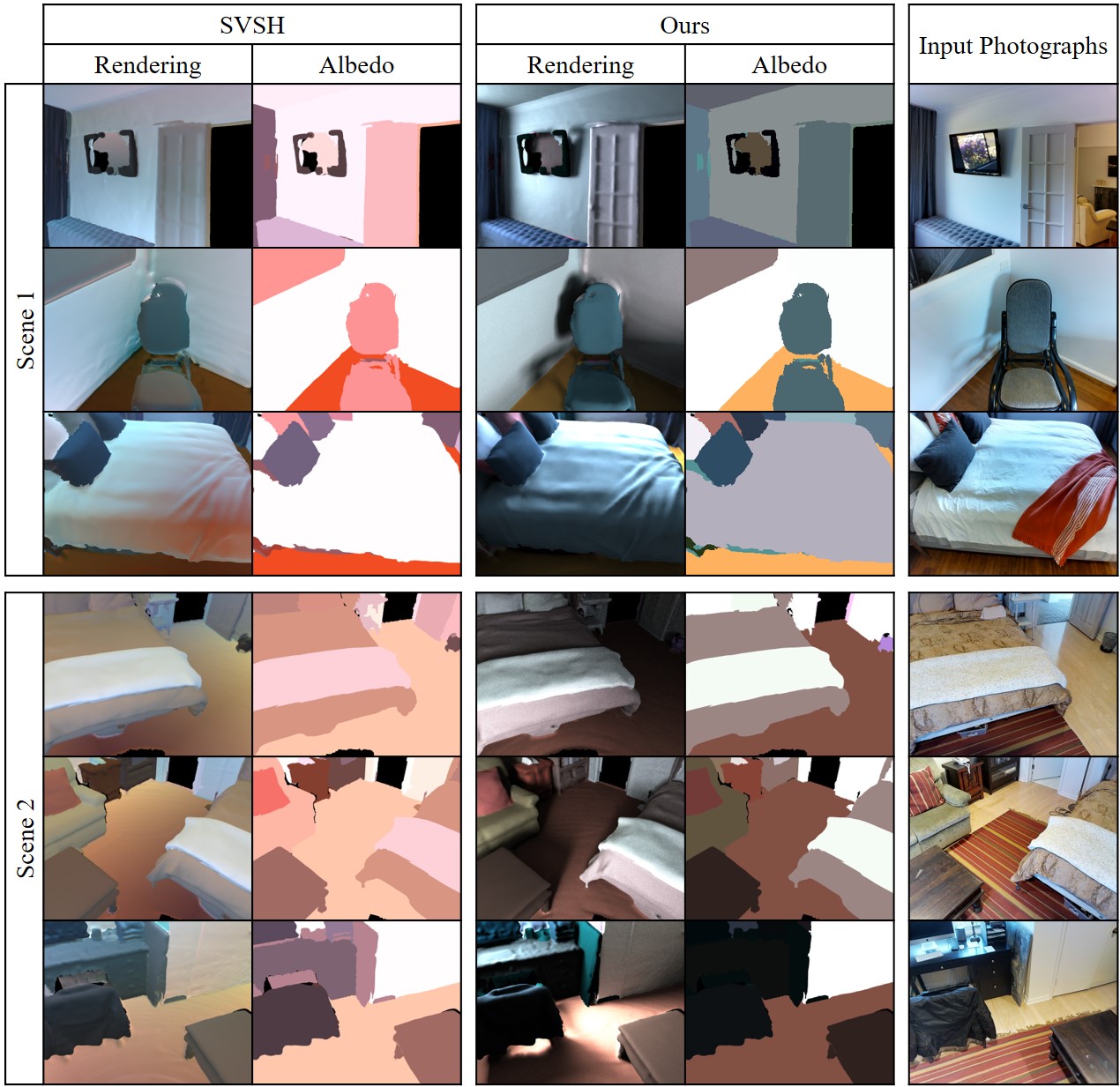}
\end{center}
\vspace{-0.5cm}
   \caption{Evaluation on real scenes: (right) input is 3D scanned geometry and photographs. We employ object instance segmentation to estimate the emission and the average albedo of every object in the scene. Our method is able to optimize for the illumination and shadows. Other methods usually do not take occlusions into account and fail to model shadows correctly. 
   Views 1 and 2 of Scene 2 show that if the light emitters are not present in the input geometry, our method gives an incorrect estimation.}
\label{fig:matterport}
\end{figure}

\paragraph{Evaluation on real data.} We use the Matterport3D~\cite{Matterport3D} dataset to evaluate our method on real captured scenes obtained through 3D reconstruction. The scene was parameterized using the segmentation provided in the dataset. Due to imperfections in the data, such as missing geometry and inaccurate surface normals, it is more challenging to perform an accurate light transport simulation. Nevertheless, our method produces impressive results for the given input. After the optimization, the optimized light direction matches the captured light direction and the rendered result closely matches the photograph. Fig.~\ref{fig:matterport} shows a comparison to the SVSH method.

The albedo of real-world objects varies across its surface. Inverse Path Tracing is able to compute an object's average albedo by employing knowledge of the scene segmentation. To reproduce fine texture, we refine the method to optimize for each individual triangle of the scene with adaptive subdivision where necessary. This is demonstrated in Fig.~\ref{fig:matterport_subdivision}.

\begin{figure}[htb!]
\begin{center}
\includegraphics[width=\linewidth]{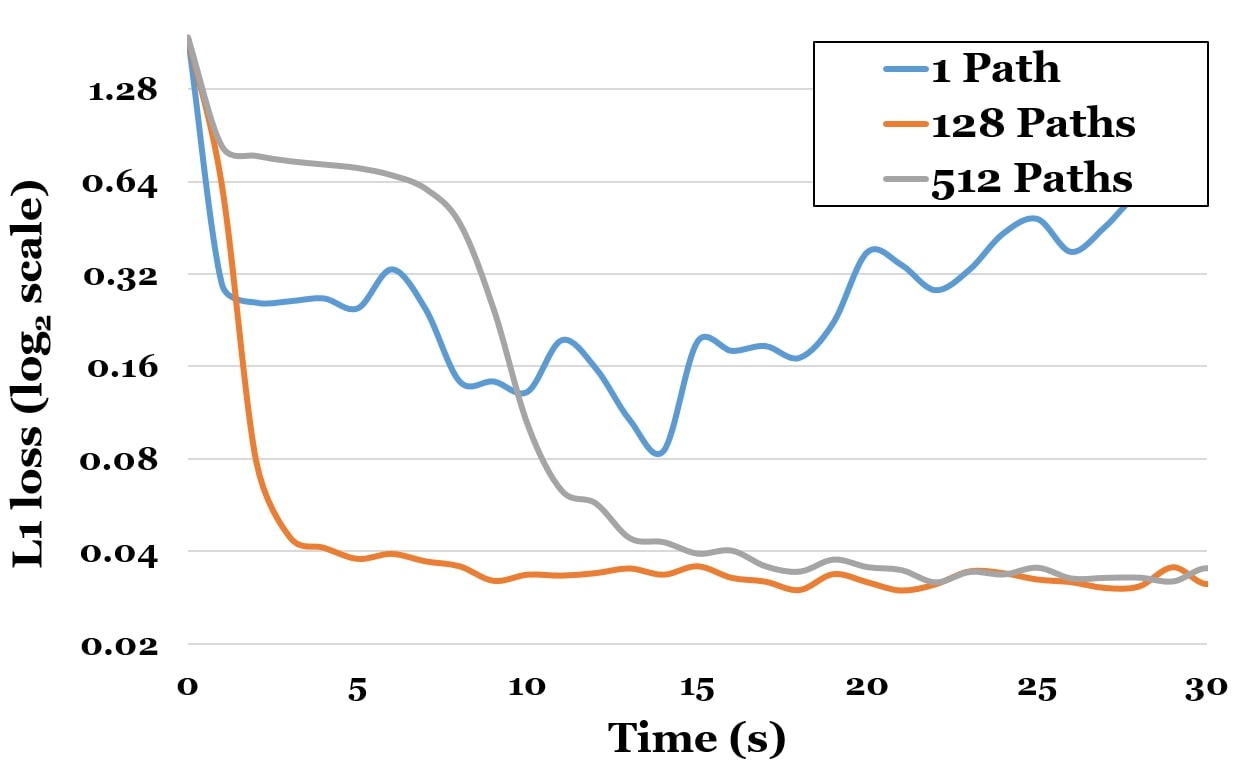}
\end{center}
\vspace{-0.6cm}
   \caption{Convergence with respect to the number of paths used to estimate the pixel color. If this is set too low, the algorithm will fail. }
\label{fig:plot_srays}
\end{figure}

\paragraph{Optimizer Ablation.}
There are several ways to reduce the variance of our optimizer. One obvious way is to use more samples to estimate the pixel color and the derivatives, but this also results in slower iterations. 
Fig.~\ref{fig:plot_srays} shows that the method does not converge if only a single path is used. A general recommendation is to use between $2^7$ and $2^{10}$ depending on the scene complexity and number of unknowns.

\begin{figure}[htb!]
\begin{center}
\includegraphics[width=\linewidth]{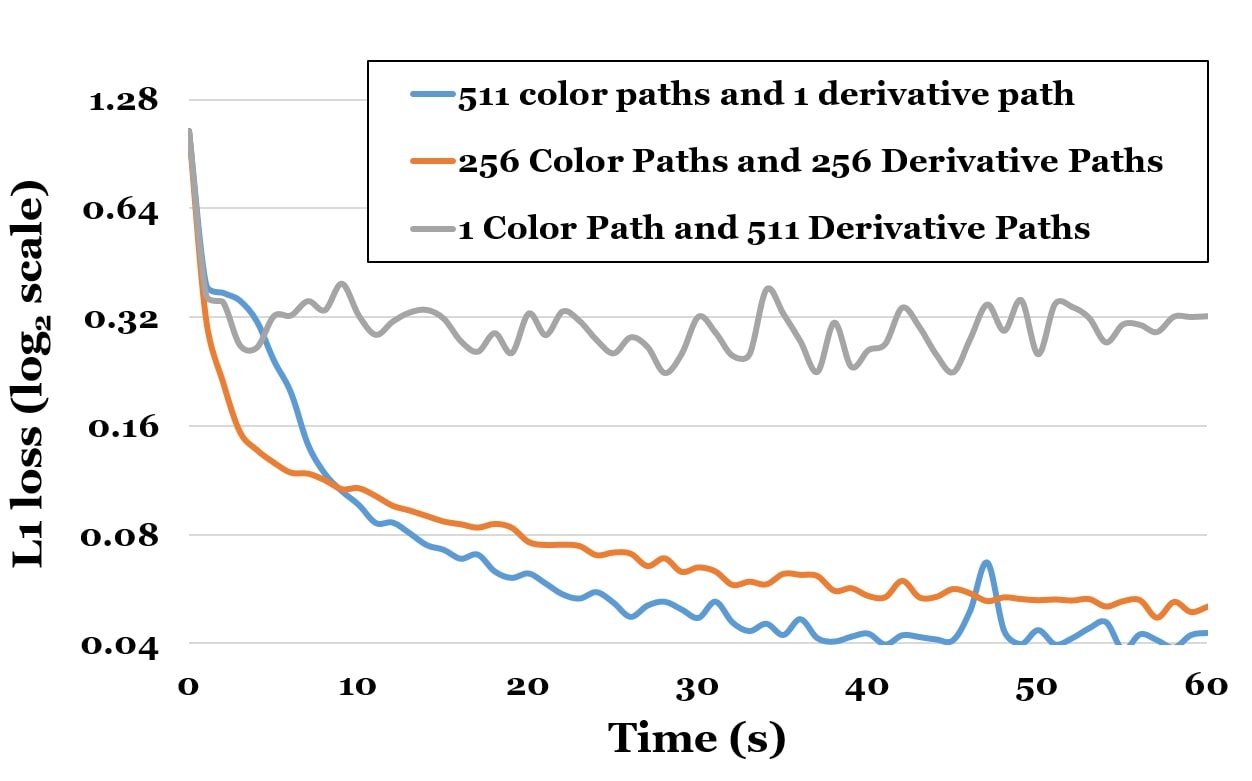}
\end{center}
\vspace{-0.6cm}
   \caption{Convergence with respect to distributing the available path samples budget between pixel color and derivatives. It is best to keep the number of paths high for pixel color estimation and low for derivative estimation.}
\label{fig:plot_samples}
\end{figure}

Another important aspect of our optimizer is the sample distribution for pixel color and derivatives estimation. Our tests in Fig.~\ref{fig:plot_samples} show that minimal variance can be achieved by using one sample to estimate the derivatives and the remaining samples in the available computational budget to estimate the pixel color.


\paragraph{Limitations.} Inverse Path Tracing assumes that high-quality geometry is available. 
However, imperfections in the recovered geometry can have big impact on the quality of material estimation as shown in Fig.~\ref{fig:matterport}. 
Our method also does not compensate for the distortions in the captured input images. 
Most cameras, however, produce artifacts such as lens flare, motion blur or radial distortion. 
Our method can potentially account for these imperfections by simulating the corresponding effects and optimize not only for the material parameters, but also for the camera parameters, which we leave for future work.

\begin{table}
\begin{center}
\begin{tabular}{|l|c|c|c|}
\hline
Method & Scene 1 & Scene 2 & Scene 3 \\
\hline\hline
SVSH Rendering Loss & 0.052 & 0.048 & 0.093 \\
Our Rendering Loss  & \textbf{0.006} & \textbf{0.010} & \textbf{0.003} \\
\hline\hline
SVSH Albedo Loss & 0.052 & 0.037 & 0.048 \\
Our Albedo Loss  & \textbf{0.002} & \textbf{0.009} & \textbf{0.010} \\
\hline
\end{tabular}
\vspace{-0.3cm}
\end{center}
\caption{Quantitative evaluation for synthetic data. We measure the L1 loss with respect to the rendering error and the estimated albedo parameters. Note that our approach achieves a significantly lower error on both metrics.}
\label{tab:quantitative}
\end{table}

\section{Conclusion}

We present Inverse Path Tracing, a novel approach for joint lighting and material estimation in 3D scenes.
We demonstrate that our differentiable Monte Carlo renderer can be efficiently integrated in a nested stochastic gradient descent optimization.
In our results, we achieve significantly higher accuracy than existing approaches.
High-fidelity reconstruction of materials and illumination is an important step for a wide range of applications such as virtual and augmented reality scenarios.
Overall, we believe that this is a flexible optimization framework for computer vision that is extensible to various scenarios, noise factors, and other imperfections of the computer vision pipeline.
We hope to inspire future work along these lines, for instance, by incorporating more complex BRDF models, joint geometric refinement and completion, and further stochastic regularizations and variance reduction techniques.

\section*{Acknowledgements}
This work is funded by Facebook Reality Labs. We also thank the TUM-IAS Rudolf M{\"o}{\ss}bauer Fellowship (Focus Group Visual Computing) for their support.
We would also like to thank Angela Dai for the video voice over and Abhimitra Meka for the LIME comparison.

{\small
\bibliographystyle{ieee}
\bibliography{main}
}

\begin{appendix}
\clearpage
\newpage
\section*{APPENDIX}
\vspace{-0.4cm}

\begin{figure}[htb!]
\begin{center}
\includegraphics[width=0.99\linewidth]{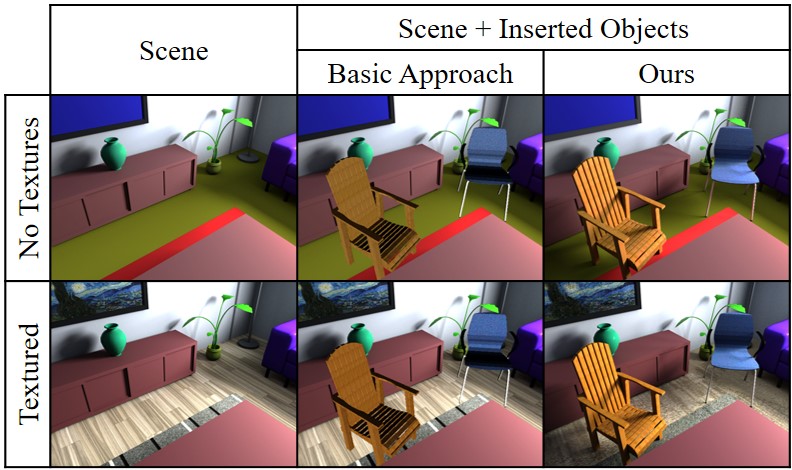}
\end{center}
\vspace{-0.5cm}
\caption{Mixed-reality setting: we insert two new 3D objects (chairs) into an existing 3D scene. Our goal is to find a consistent lighting between the existing and newly-inserted content. In the middle column, we show a naive compositing approach; on the right the results of our approach. The naive approach does not take the 3D scene and light transport into consideration, and fails to photo-realistically render the chair.}
\label{fig:synth_chairs}
\end{figure}

In this appendix, we provide additional quantitative evaluations of our design choices in Sec.~\ref{sec:supp_ablations}. 
To this end, we evaluate the choice of the batch size, the impact of the variance reduction, and the number of bounces for the inverse path tracing optimization.
In addition, we provide additional results on scenes with textures, where we evaluate our subdivision scheme for high-resolution surface material parameter optimization; see Sec.~\ref{sec:supp_textured}.
Sec.~\ref{sec:supp_lime} presents a quantitative comparison to another material estimation method.
In Sec.~\ref{sec:supp_mixedreality}, we provide examples for mixed-reality application settings where we insert new virtual objects into existing scenes.
Here, the idea is to leverage our optimization results for lighting and materials in order to obtain a consistent compositing for AR applications.
Finally, we discuss additional implementation details in Sec.~\ref{sec:supp_implementation_details}.

\section{Qualitative Evaluation of Design Choices}
\label{sec:supp_ablations}

\subsection{Choice of Batch Size}

\begin{figure}[htb!]
\begin{center}
\vspace{0.6cm}
\includegraphics[width=\linewidth]{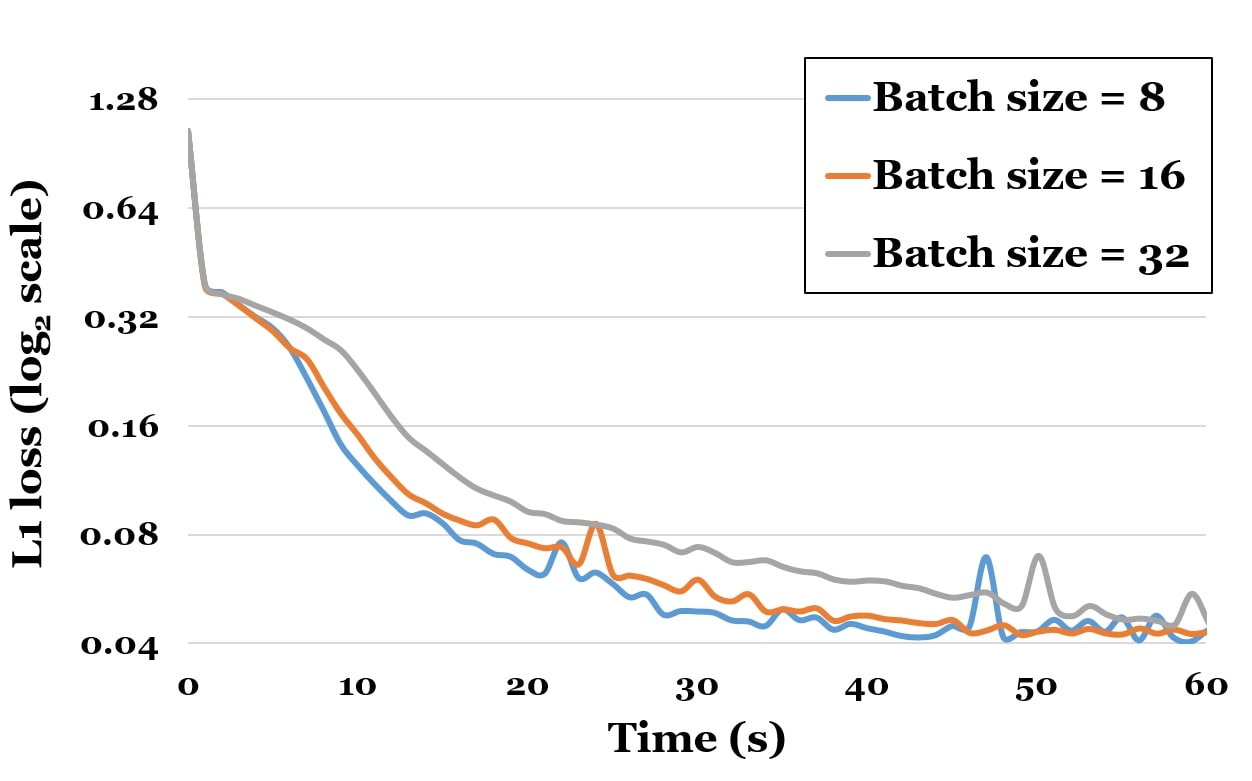}
\end{center}
\vspace{-0.5cm}
   \caption{Convergence with respect to the batch size: in this experiment, we assume the same compute/time budget for all experiments ($x$-axis), but we use different distributions of rays within each batch; \ie, we try different batch sizes.}
\label{fig:plot_batch}
\end{figure}

In Fig.~\ref{fig:plot_batch}, we evaluate the choice of the batch size for the optimization.
To this end, we assume the compute budget for all experiments, and plot the results with respect to time on the $x$-axis and the $\ell_1$ loss of our problem (log scale) on the $y$-axis. 
If the batch size is too low (blue curve), then the estimated gradients are noisy, which leads to a slower convergence; if batches are too large (gray curve), then we require too many rays for each gradient step, which would be used instead to perform multiple gradient update steps.

\subsection{Variance Reduction}
\label{ssec:varred}
\begin{figure}[htb!]
\begin{center}
\vspace{-0.3cm}
\includegraphics[width=\linewidth]{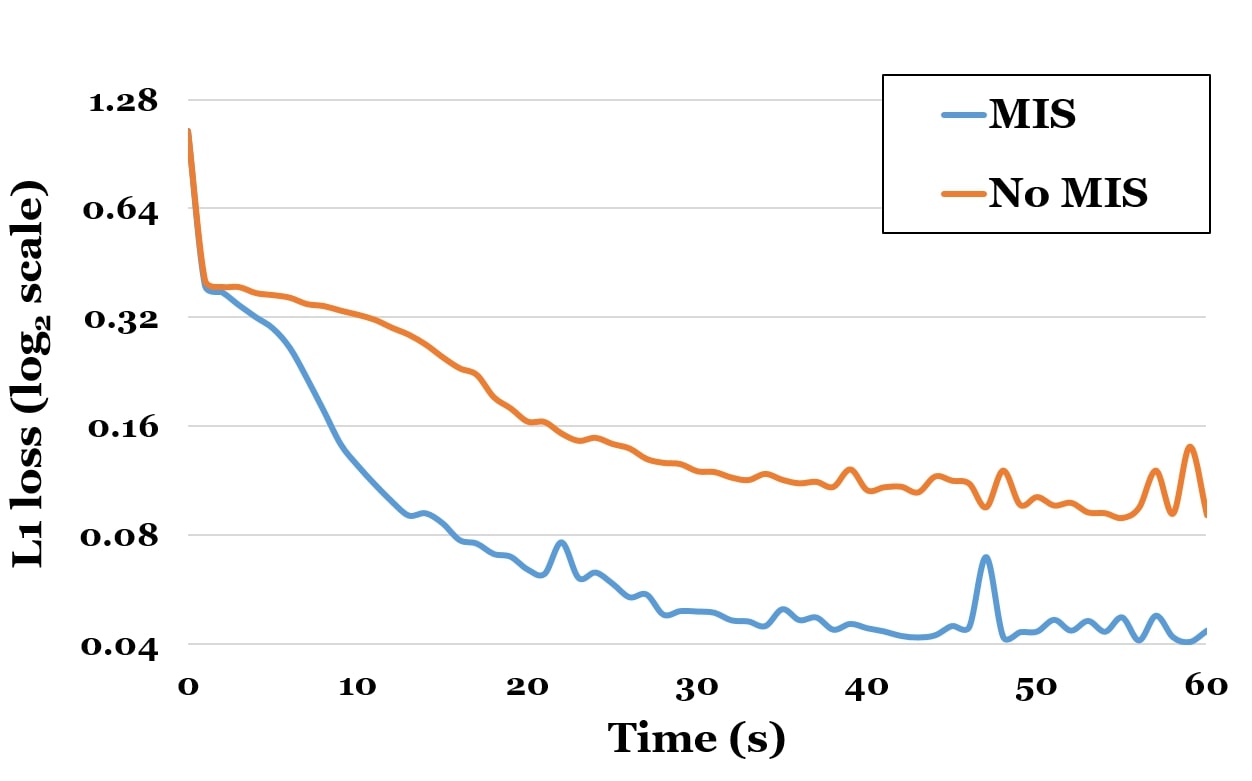}
\end{center}
\vspace{-0.5cm}
   \caption{Use of Multiple Importance Sampling during path tracing significantly improves the convergence rate.}
\label{fig:plot_mis}
\end{figure}
In order to speed up the convergence of our algorithm, we must aim to reduce the variance of the gradients as much as possible.
There are two sources of variance: the Monte Carlo integration in path tracing and the SGD, since we path trace only a small fraction of captured pixels in every batch.

As mentioned in the main paper, the gradients of the rendering integral have similar structure to the original integral, therefore we employ the same importance sampling strategy as in usual path tracing. The path tracing variance is reduced using Multiple Importance Sampling (\ie, we combine BRDF sampling with explicit light sampling)~\cite{Veach98}. 
We follow the same computation for estimating the gradients with respect to our unknowns. 
A comparison between implementation with and without MIS is shown in Fig.~\ref{fig:plot_mis}.

\subsection{Number of Bounces}

\begin{figure}[htb!]
\begin{center}
\includegraphics[width=\linewidth]{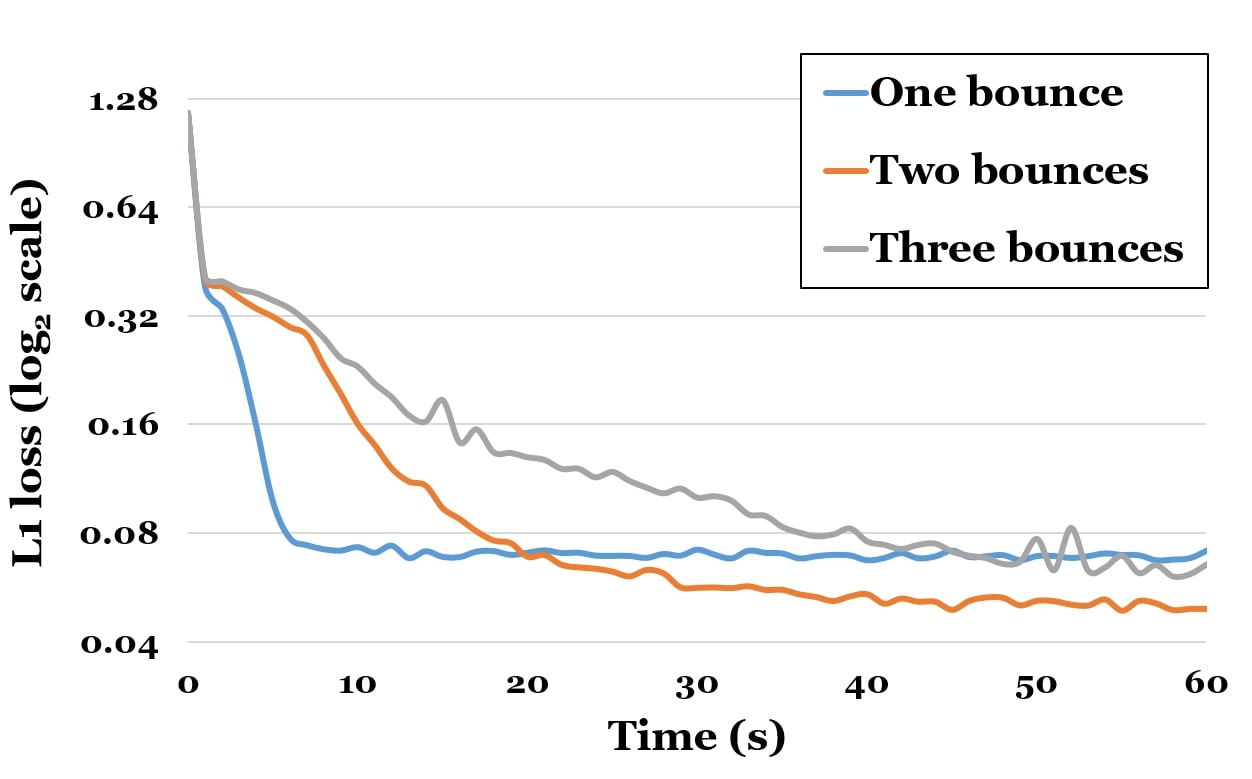}
\end{center}
\vspace{-0.6cm}
   \caption{A scene rendered with 10 bounces of light is given as input to our algorithm. We estimate emission and material parameters by using one, two, and three bounces during optimization. Two bounces are enough to capture most of the diffuse indirect illumination in the scene.}
\label{fig:plot_bounces}
\end{figure}

We argue that most diffuse global illumination effects can be approximated by as few as two bounces of light. 
To this end, we render an image with $10$ bounces and use it as ground truth for our optimization.
We try to approximate the ground truth by renderings with one, two, and three bounces, respectively (see Fig.~\ref{fig:plot_bounces}).
One bounce corresponds to direct illumination; adding more bounces allows us to take into account indirect illumination as well. 
Optimization with only a single bounce is the fastest, but the error remains high even after convergence. 
Having more than two bounces leads to high variance and takes a lot of time to converge. 
Using two bounces strikes the balance between convergence speed and accuracy.

\vspace{-0.2cm}
\section{Results on Scenes with Textures}
\label{sec:supp_textured}

\begin{figure*}[htb!]
\begin{center}
\includegraphics[width=0.99\linewidth]{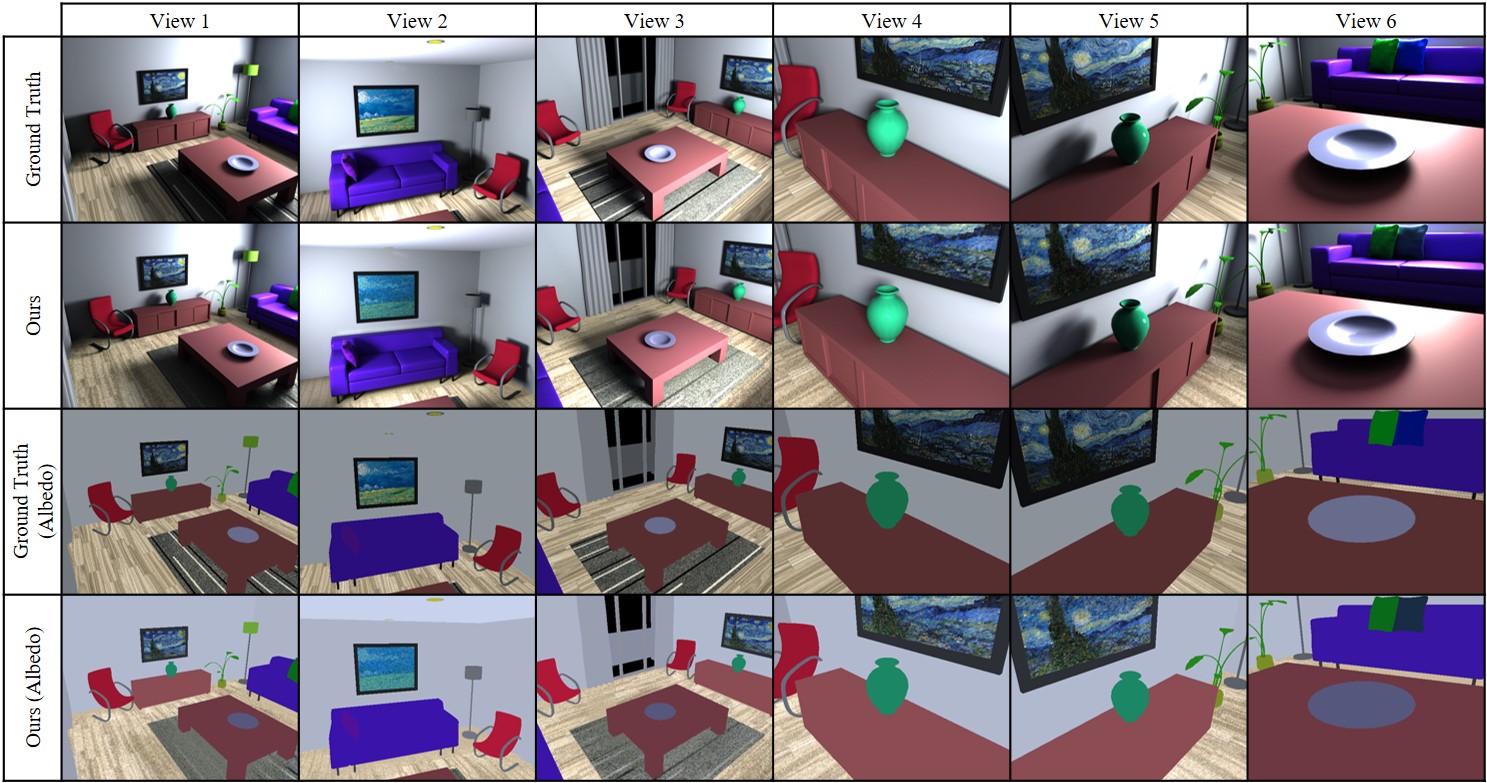}
\end{center}
\vspace{-0.2cm}
   \caption{Results of our approach on synthetic scenes with textured objects. Our optimization is able to recover the scene lighting in addition to high-resolution surface texture material parameters.}
\label{fig:synth_textures}
\end{figure*}

\begin{figure*}[htb!]
\begin{center}
\includegraphics[width=0.99\linewidth]{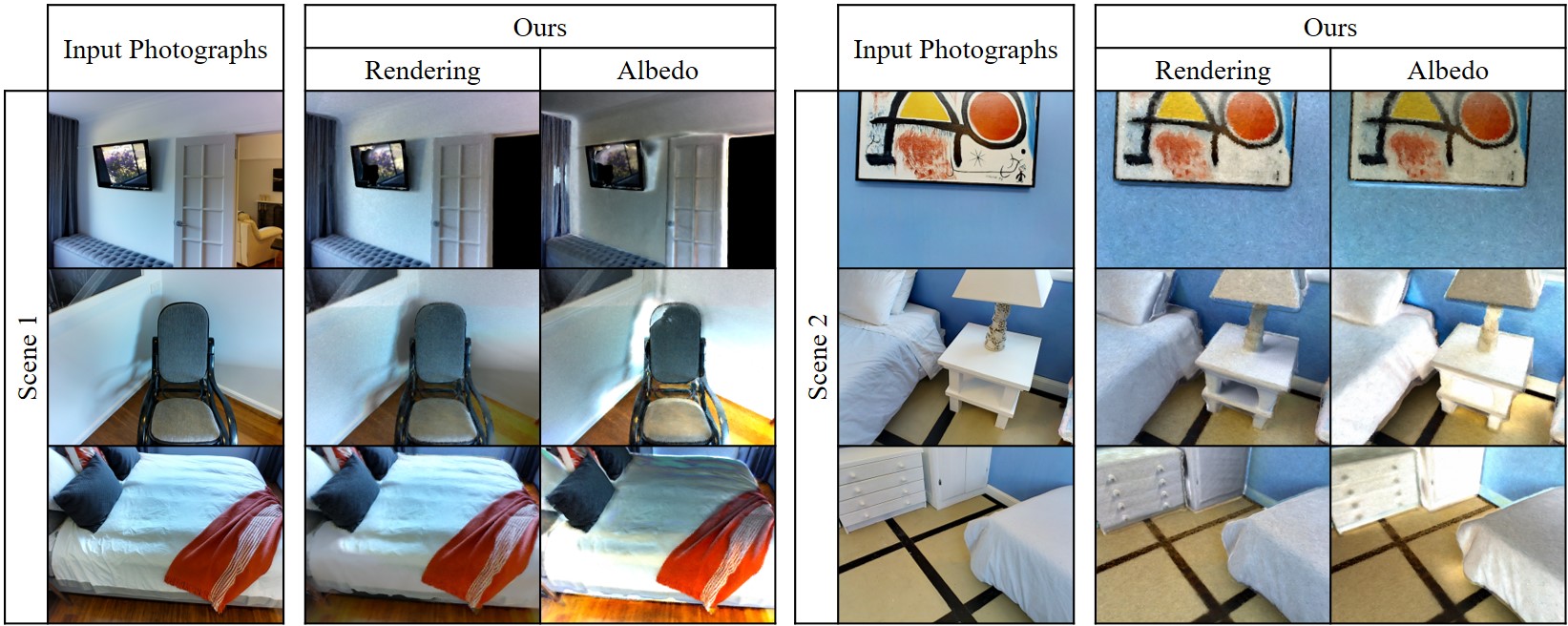}
\end{center}
\vspace{-0.2cm}
\caption{Examples from Matterport3D~\cite{Matterport3D} (real-world RGB-D scanning data) where we reconstruct emission parameters, as well as high-resolution surface texture material parameters. We are able to reconstruct fine texture detail by subdividing the geometry mesh and optimizing on individual triangle parameters. Since not all light sources are present in the reconstructed geometry, some inaccuracies are introduced into our material reconstruction. Albedo in shadow regions can be overestimated to compensate for missing illumination (visible behind the chair in Scene 1), specular effects can be baked into the albedo (reflection of flowers on the TV) or color may be projected onto the incorrect geometry (part of the chair is missing, so its color is projected onto the floor and wall).}
\label{fig:matterport_fine}
\end{figure*}

In order to evaluate surfaces with high-frequency surface signal, we consider both real and synthetic scenes with textured objects.
To this end, we optimize first for the light sources and material parameters on the coarse per-object resolution. 
Once converged, we keep the light sources fixed, and we subdivide all other regions based on the surface texture where the re-rendering error is high; \ie, we subdivide every triangle based on the average $\ell_2$ error of the pixels it covers, and continue until convergence.
This coarse-to-fine strategy allows us to first separate out material and lighting in the more well-conditioned setting; in the second step, we then obtain high-resolution material information.
Results on synthetic data~\cite{handa:etal:ICRA2014} are shown in Fig.~\ref{fig:synth_textures}, and results on real scenes from Matterport3D ~\cite{Matterport3D} are shown in Fig.~\ref{fig:matterport_fine}.

\section{Additional Comparison to Data-driven Approaches}
\label{sec:supp_lime}
We compare our approach to Meka et al.~\cite{Meka:2018:LLI} and present quantitative results in Tab.~\ref{tab:lime}. 
Please note that our approach is not limited to a single material of a single object at a time. 
The other data-driven references are mostly on planar surfaces only and/or assume specific lighting conditions, such as a single point light close to the surface.
 
\begin{table}
\begin{center}
\begin{tabular}{|l|c|c|}
\hline
Method & LIME~\cite{Meka:2018:LLI} & Ours \\
\hline\hline
Object 1 & 0.45\% & \textbf{0.00037\%} \\
Object 2 & 1.37\% & \textbf{0.14\%} \\
\hline
\end{tabular}
\vspace{-0.6cm}
\end{center}
\caption{We compare the relative error between the estimated diffuse albedo for two objects. We outperform LIME even though our method is not restricted to the estimation of only a single material at a time.}
\vspace{-0.3cm}
\label{tab:lime}
\end{table}

\section{Object Insertion in Mixed-reality Settings}
\label{sec:supp_mixedreality}

One of the primary target applications of our method is insertion of virtual objects into an existing scene while maintaining a coherent appearance.
Here, the idea is to first estimate the lighting and material parameters of a given 3D scene or 3D reconstruction.
We then insert a new 3D object into the environment, and re-render the scene using both the estimated lighting and material parameters for the already existing content, and the known intrinsics parameters for the newly-inserted object.
A complete 3D knowledge is required to produce photorealistic results, in order to take interreflection and shadow between objects into consideration.

In Fig.~\ref{fig:synth_chairs}, we show an example on a synthetic scene where we virtually inserted two new chairs.
As a baseline, we consider a naive image compositing approach where the new object is first lit by spherical harmonics lighting and then inserted while not considering the rest of the scene; this is similar to most existing AR applications on mobile devices.
We can see that a naive compositing approach (middle) is unable to produce a consistent result, and the two inserted chairs appear somewhat out of place.
Using our approach, we can estimate the lighting and material parameters of the original scene, composite the scene in 3D, and then re-render. 
We are able to show that we can produce consistent results for both textured and non-textured optimization results (right column).

In Fig.~\ref{fig:matterport_teddy}, we show a real-world example on the Matterport3D~\cite{Matterport3D} dataset, where we insert a virtual teddy into the environment. 
To this end, we first estimate lighting and surface materials in a 3D scan; we then insert a new virtual object, render it, and then apply the delta image to the original input. 
Compared to the SVSH baseline, our approach achieves significantly better compositing results.

\section{Implementation Details}
\label{sec:supp_implementation_details}

We implement our inverse path tracer in C++, and all of our experiments run on an 8-core CPU. 
We use Embree~\cite{Wald:2014:EKF} for the ray casting operations. 
For efficient implementation, instead of employing automatic differentiation libraries, the light path gradients are computed using manually-derived derivatives.

We use ADAM \cite{Kingma:2015:AMS} as our optimizer of choice with an initial learning rate of $5\cdot10^{-3}$. 
We further use an initial batch size of 8 pixels which are uniformly sampled from the set of all pixels of all images. 
We found marginal benefit of having larger batches, but we believe there is high potential in investigating better sampling strategies.
In all our experiments, the emission and albedo parameters are initialized to zero.

For every pixel in the batch, we need to compute an estimate of the pixel color based on the current value of the unknown material and emission parameters. 
This estimated color is compared against the ground truth color and a gradient is computed depending on the choice of the loss function. 
For most commonly used loss functions, this gradient will involve a multiplication of the estimated pixel color and its derivative with respect to the unknown parameters. 
Since these are random variables (approximated by Monte Carlo integration), it is important that they are calculated from independent samples to avoid bias. 
We use path tracing with multiple importance sampling for the computation of the pixel color, but any unbiased light transport method will produce the correct result.

We extend our path tracer to analytically compute derivatives \wrt emission and materials parameters as defined by Eq. \ref{eq:emission_gradients} and \ref{eq:material_gradients}. 
To this end, we pass a reference to a structure holding the derivatives to our ray casting function. The product of BSDFs in Eq. \ref{eq:emission_gradients} is incrementally calculated at each bounce. 
Given that $L_e(x_i)$ is the unknown emission parameter on surface $i$, the derivative \wrt this emission parameter is equal to the product of the BSDFs at each surface intersection from surface $i$ to the sensor. 
The derivatives \wrt to the materials are computed in similar manner. 
As per chain rule, we multiply the throughput by the derivative of the BSDF \wrt the unknown material parameters to obtain the derivative of the pixel color \wrt the unknown material parameters.

We implement multiple importance sampling, a combination of light source sampling and BRDF importance sampling.
The importance for light source sampling is based on the unknown emission parameters which may change in every iteration of our optimization. 
An efficient data structure is needed to store the sampling probabilities for every object. 
We implement a binary indexed tree (also known as Fenwick tree) for this purpose. 
This provides logarithmic complexity for both reading and updating the probabilities.

Finally, to make the optimization more robust, we propose a coarse-to-fine approach, where we first optimize for one emission and one material parameter per object instance. 
Most scenes have only a few emitters, so we employ an L1-regularizer on all the emission parameters. After convergence, the result is refined by optimizing for material parameters of individual object triangles. 
The light sources stay fixed in this phase, but their emission value may still change. 
In the end, the triangles may be subdivided as explained in Sec.~\ref{sec:supp_textured} to further improve the results.

\end{appendix}

\end{document}